# Text classification in shipping industry using unsupervised models and Transformer based supervised models

Ying Xie (School of Management, Cranfield University, Bedford, UK)
Ying.xie@cranfield.ac.uk
Dongping Song (School of Management, University of Liverpool, UK)
Dongping.Song@liverpool.ac.uk

Obtaining labelled data in a particular context could be expensive and time consuming. Although different algorithms, including unsupervised learning, semi-supervised learning, self-learning have been adopted, the performance of text classification varies with context. Given the lack of labelled dataset, we proposed a novel and simple unsupervised text classification model to classify cargo content in international shipping industry using the Standard International Trade Classification (SITC) codes. Our method stems from representing words using pretrained Glove Word Embeddings and finding the most likely label using Cosine Similarity.

To compare unsupervised text classification model with supervised classification, we also applied several Transformer models to classify cargo content. Due to lack of training data, the SITC numerical codes and the corresponding textual descriptions were used as training data. A small number of manually labelled cargo content data was used to evaluate the classification performances of the unsupervised classification and the Transformer based supervised classification. The comparison reveals that unsupervised classification significantly outperforms Transformer based supervised classification even after increasing the size of the training dataset by 30%. Lacking training data is a key bottleneck that prohibits deep learning models (such as Transformers) from successful practical applications. Unsupervised classification can provide an alternative efficient and effective method to classify text when there is scarce training data.

## I. INTRODUCTION

Alongside the rapid development of digital products and technologies, container shipping services and operations are being transformed into a digital era, facilitated by applying artificial intelligence (including machine learning, deep learning, natural language processing, etc.), robotics, 5G, blockchain and Internet of Things sensors. Past research has developed machine learning models for demand prediction, quayside operation, vehicle trajectories, ship fuel consumption and on time performance prediction. In summary, the applications of data analytics and machine learning in container ports are predominantly on port throughput forecasting and vessel estimated time of arrival with very few on port operations management. This paper proposes a Natural Language Processing (NLP) based technique to effectively classify container content. Container content is an important feature used in machine learning models to predict container destinations in port operations management. Classifying container content into the Standard International Trade Classification (SITC) codes reduces the dimensionality of container content, which improves the interpretability of the machine learning models but without scarifying prediction accuracy.

In real world applications, labelled text data are often scarce, and obtaining labelled data could be labour intense and costly. Although different algorithms, including traditional supervised text classification, unsupervised classification, semi-supervised classifications, deep learning models have developed, the performance of text classification varies with context. Recent advancement in deep learning models leads to state-of-the-art results in many NLP tasks, including text classification, question answering, translation, text generation and many more. Transformers are a class of deep neural network models that significantly improve performances on a wide array of standard NLP tasks, by leveraging large-scale pre-training, transfer learning and parallel processing of the words in the documents. In this research, several Transformer models were applied to classify cargo content in international shipping industry in a supervised manner. Due to lack of training data, the Standard International Trade Classification (SITC) codes and the corresponding descriptions were used as training data, but the number of training data is very small. On the other hand, we proposed an innovative and simple unsupervised classification model which represents texts using Glove Embeddings and assigns classification label using Cosine Similarity.

## II. TEXT CLASSIFICAITON

### A. Text Representation

In text classification, the most fundamental step is text representation [1], i.e., a process of converting unstructured text documents to numerical vectors, to make them mathematically computable. The most widely applied text representation method is the bag of words (BoW) [2] and Term Frequency-Inverse Document Frequency (TF-IDF). BoW maps texts into feature vectors according to the occurrences of the words (i.e., features) in that text. TF-IDF evaluates how relevant a word is to a document in a collection of documents.

The main shortfall of BoW and TF-IDF is the inability of capturing context information of words or texts. This is where word embeddings were developed. Word embeddings model words in vectors of real numbers and capture syntactic features and semantic word relationships [3]. Word embeddings are commonly used in deep learning models and past research showed that the performance of text classification largely depends on effectiveness of the word embeddings [4]. Word embeddings can be either trained using the embedding layer in a deep learning model, alternatively, pre trained word embeddings can be adopted as an initialiser in the embedding layer, such as Google's Word2Vec, BERT, Standford's GloVe FastText developed by Facebook. Word2Vec FastText and GloVe are context independent word embeddings that models one vector for each word by integrating all the different senses of the word into one vector. BERT produces different word embeddings for a word by capturing its context and considering word order and position in a text. One group of studies built text representations by aggregating the word embeddings as document embeddings

[4][5], while the other group built text representations by jointly learning word/document and document embeddings [6]; the text representations were then fed into a classifier.

## B. Deep Learning-Based Text Classification

- *ANN, RNN and CNN*

  Extracting features manually from input data is extremely time-consuming, and it requires strong knowledge of the subject as well as the domain. To overcome this problem, deep learning models are developed to automate the process of feature engineering and selection. The ability of automatically selecting features makes deep learning models popular choices for text classification tasks [7]. Three representative deep networks are Artificial Neural Network (ANN) [4][8], Convolutional Neural Network (CNN) [9][10] and Recurrent Neural Network (RNN). ANN use Bag of Words to represent text, and the vector representations are fed into one or more feed-forward layers to perform classifications [8]. Liu et al. [11] and Luo et al. [12] used LSTM, a specific type of RNN, to learn text representation. Kim[10] used CNN for sentence classification, while Conneau et al. [13] applied CNNs at character level and achieved promising results.

- *Attention Mechanism*

  To further increase the representation flexibility of such models and better capture semantics and words relationships, researchers introduced various mechanisms to Neural Network based models. Attention mechanisms have been introduced as an integral part of models to model word associations [14]. Yang et al. [14] applied a hierarchical two-level attention network at the word and sentence level, for text classification. This model outperforms substantially on text classification problems.

  Zhou et al. [15] applied the hierarchical attention model to multi-lingual sentiment classification. The sentence-level attention model learns importance of sentences in a document, while the word level attention model learns the importance of words in each sentence, then the overall sentiment is decided.

- *Transformers*

  The computational costs to hand sequential input data in RNNs, or to capture relationships between words in CNNs increase potentially along with the length of the sentence. To overcome this limitation, Transformers were developed to compute an attention score for every word in a sentence or document in parallel, and the attention score is used to model the influence each has on another [16].

  Transformers use deep neural network architectures (for example, a neural network architecture with 12 or 24 layers) and are pre-trained on much larger amounts of text corpora to learn contextual text representations. Transformers have been trained on a large amount of text in a self-learning manner. Self-learning is a type of training where the model trains itself by leveraging one part of the data to predict the other party, and to generate labels as the learning progresses, hence it eliminates the necessity of data labelling and reduces cost and time of manual labelling [17]. One of the most widely used Transformers is Bidirectional Encoder Representations for Transformers (BERT) [18]. BERT was pretrained from unlabelled data extracted from English Wikipedia and BooksCorpus, on two tasks: masked language modelling and next sentence prediction. As a result of pretraining, it require less resources for task specific fine tuning when it is applied to create text representations [18][19]. Applying deep pretrained Transformers models to downstream NLP tasks has achieved state-of-the-art performance, including text classification, albeit with scarce label sets [20][21].

## C. Multi Class Text Classification

Most researchers used open-source text classification datasets to train the deep learning models, and to conduct experiments and analyse classification performance. These datasets include: (1) sentiment analysis datasets, such as Yelp (185), IMDB (186), Standford Sentiment Treebank 43, Amazon product reviews (189); (2) News Classification datasets, such as 20 Newsgroups 190, Reuters News 191, (3) Topic Classification datasets, such as DBpedia (195), Ohsumed medicine database (196), ERU-lex EU Law database 197, Web of Science (WOS) dataset (136); (4) Natural Language Inference dataset, such as Stanford NLI (208), Mutli-Genre NLI (209, etc. Most of these datasets have 2 -5 labels, such as the IMDb with 2 binary classes, AGNews has 4 categories, and Yell news has 5 fine grained sentiment labels. The ERU-lex European Union law has 19,314 documents and 3,956 categories. Past research has shown the difficulty in performing large scale text classification increases significantly when the number of class labels exceeds 100 [22].

Different from the past research that uses public open-source data to test the classification models, this research examines how to classify shipping cargo content using 10 category labels. The class labels are mutually exclusive, and one cargo content is associated with only one label. The absence of labelled training data makes the classification task very challenging. We propose a novel and simple model for unsupervised text classification and evaluation. Our method stems from representing words using pretrained Glove Word Embeddings and finding the most likely label using Cosine Similarity.

To make a comparison, a second model is built as flat supervised classification and it is trained using SITC textual descriptions and manually labelled data. Three Hugging Face Pre-trained Transformers [16], including BERT, RoBERTa and XLNet, are used to represent text and perform supervised classification. Hugging Face provides pre-trained models for a variety of transformer architectures and these models have shown promise in improving results on tasks with a small amount of labelled data, which is the challenge facing by this research. A comparison of transformer-based supervised classification and Cosine similarity based unsupervised text classification is also made. Unsupervised text classification proves to be effective in performing the multi class text classification, with higher accuracy score.

### III. THE NEEDS TO CLASSIFY CARGO CONTENT

Within the hundreds of millions of containers received by a shipping port, many of them contain the same content but have slightly content names. That means some cargo content is unique in names, but they belong to the same content category, for example, "HOBBY RACK BOX1" and "HOBBY RACK BOX2" belong to the same category. Therefore, when using machine learning models to analyse the operation of containers, such as the destinations of containers, or the haulage companies appointed to transport containers, it is necessary to classify the cargo content into the same category to reduce the feature dimensionality, to remove redundant information and noise, and to reduce computing time in machine learning models.

Manually labelling a higher number of text documents can be a very labour intensive and intellectually challenging activity; hence, the labelled training documents are often unavailable. Even when subject knowledge is applied to label text documents, variation could be high, especially in multi-class classification where there are many labels. The variation in classification is caused by personal opinions and is open to different interpretations [23]. For these reasons, the labelled cargo content does not exist. In the absence of labelled training data, we proposed an unsupervised classification to categorise the container content.

## IV. Proposed Unsupervised Classification

The text document is a collection of unique shipping cargo content which is retrieved from a shipping port's operation terminal. The 10 category labels are derived from the Level 1 SITC codes. SITC is a product classification of the United Nations used for international trade. SITC has five levels of classification structures, comprising Level 1 that defines goods in 10 Sections using one-digit code (0-9), Level 2 that defines goods in 67 divisions using two-digit codes, Level 3 with 261 groups defined using three-digit codes, Level 4 that defines 1033 subgroups using four-digit codes and Level 5 that defines goods in 2970 items using five-digit codes. An example of the hierarchical SITC codes is given in Table 1:

TABLE I. An example of hierarchical Standard International Trade Classification codes

| Level | Code | Textual Description |
|---|---|---|
| 1 | 0 | Food and live animals |
| 2 | 00 | Live animals |
| 3 | 001 | Live animals other than animals of division 03 |
| 4 | 0011 | Bovine animals, live |
| 5 | 00111 | Pure-bred breeding animals |

TF-IDF vectorizer was applied as weights to Word Embedding to create Weighted Embedding Matrix. Some of the cargo content has detailed description, such as "FROZEN INDIAN MACKEREL WHOLE ROUND", "BTM MACKEREL IN TOMATO SAUCE", "PREPARED OR PRESERVED TUNAS, S", and "CANNED TUNA", while some of the content is brief and unspecific, such as "FURNITURE". To maximise the classification accuracy, we built category word embeddings of the most detailed SITC Level 5 structure to calculate Cosine similarity with the text word embeddings.

### A. Unsupervised classification Using SITC

The multiple class text classification was applied to label similar container content using the predefined Level 5 SITC codes. The aim of classification is to systematically and objectively transforming the large number of individual content items into meaningful categories, reducing the dimension size of this feature and saving computing time.

*1) Pre-processing categorical varables using Natural Language Processing*

A pipeline of NLP steps was performed to pre-process the text data contained:

*a) Preprocessing using Regular Expresssions (Regex):* Regex is used to remove non-alphanumeric characters from the text and to remove extra spaces and punctuations. Regex is also performed to decontract words (replacing the words like can't with cannot) and to remove newlines and tabs (such as /, -, /n). All the texts were converted to lower case to avoid algorithms interpreting same words with different cases as different.

*b) Removing stop words, tokenisation and lemmatization:* Stopwords are commonly used words that do not carry much meaning or weight compared with other words, such as "the", "and", "or", etc. Tokenisation is the process of splitting a string into individual tokens. A sentence can be tokenised into n-grams tokens (n=1, 2, 3…), i.e., a sequence of n words. Lemmatisation is the process of reducing the number of words into a single word by combining common words together, for example, "blogging" and "blogger" are lemmatised into the root word "blog".

*c) Feature extraction*: text data needs to be covered into numerical vectors before being processed by any machine learning algorithm. Word Embeddings map words from the vocabulary to low dimensional, dense and real valued vectors, by capturing syntactic and semantic information. Words from the same context are clustered together and are represented by similar vectors. Each dimension of the vectors represents a different aspect of words [24]. For example, a word embedding scheme W maps a word to a d-dimensional vector: word$\rightarrow R^d$, where d is usually an integer of 100, 200, 300 or 500. For example, the words "machine" and "learning" can be represented as:

W('machine')=(-0.406, 0.306, -0.012, …)

W('learning)=(-0.591, 0.432, 0.721,…)

In machine learning, embeddings not only aid deep learning but also help feature pre-processing [25]. The commonly used pre-trained word embeddings techniques are Google's Word2Vec [26]. Stanford's GloVe [27] and Facebook's Fasttext [4]. In this research, a pre-trained Glove embedding "glove.6B.100d.txt" was used to produce vector representation of words in general cargo content and the SITC codes. The pre-trained Glove embedding maps a word to a 100-dimensional vector: word$\rightarrow R^{100}$. Each cargo content and SITC code is represented as a N x100 matrix, where N is the number of words in the cargo content or the SITC code.

*2) Integrating unsupervised learning and classification*

Text similarity between two pieces of text is determined by lexical similarity (closeness in words) and meaning (semantic similarity). Jia et al. [28] used unsupervised pre-training to estimate an embedding layer and represent categorical variables. In a novel unsupervised text classification model, elements in TF-IDF matrix were applied as weights to Word Embedding to create Weighted Embedding Matrix. Weighted Word embedding proved to improve text classification accuracy, in comparison with non-weighted word embedding models [29].

When comparing similarity between cargo content and the SITC codes, we need to determine the text similarity, i.e., how close two pieces of text are both in lexical similarity and semantic similarity. A commonly used similarity measure for text clustering is Cosine Similarity [30]. Cosine similarity measures the similarity between vectors using the cosine value of the angle and determines whether two vectors are pointing in roughly the same direction. Compared with Euclidean distance which directly measures the linear interval or length between vectors, Cosine similarity pays more attention to the difference between the relative levels of the dimensions, hence, it has a better effect than Euclidean distance. Even if the two texts are distant by the Euclidean distance, they may still be oriented together in terms of Cosine similarity. The smaller the angle, higher the cosine similarity.

Assuming there are two nonzero vectors in space x and y, the Cosine similarity is defined as the Cosine of the angle between the two vectors:

Cosine similarity($|x \cdot y|$)=xy/( $\| x \|$  $\| y \|$ )  (1)

The following Algorithm 1 was executed to classify container content into predefined categories:

| Algorithm 1: The algorithm to perform unsupervised classification |  |
|---|---|
| 1 | **Input:** unique cargo content and a list of predefined categories of SITC codes |
| 2 | **Output:** unique cargo content labelled with a predefined SITC code |
| 3 | Append the cargo content to the predefined categories of SITC codes and constitute a corpus C $_{m \times n}$ where m denotes the number of documents in C and n is the number of unique terms in C |
| 4 | Obtain a TI-IDF term document matrix M with C, M $\in R^{m \times n}$ |
| 5 | For each document t in C, pad it to length n, t $\in \{1, \cdots, m\}$ |
| 6 | Get the word embedding matrix E $\in R^{n \times l}$, where l is the embedding size of the pre-trained vectors created by GloVe model (l=100 in this research) |
| 7 | Create an empty weighted document embedding matrix W $\in R^{m \times l}$ |
| 8 | **For each** t $\in \{0, \cdots, m-1\}$ do: |
| 9 | **For each** j $\in \{0, \cdots, n-1\}$ do: |
| 10 | W[t, j+1] = W[t, j]+ E[j]∘M[t][j], where ∘ denotes the element wise multiplication of the two matrices |
| 11 | **End for** |
| 12 | **End for** |
| 13 | Calculate the pair wise Cosine similarity of the W |
| 14 | Find the index SITC $\in \{1, \cdots, m-1\}$ that achieves the maximum similarity between W[SITC] and W[m] |
| 15 | Assign SITC to the cargo content as the classification label |
| 16 | Record the classification label SITC in the Dataframe DF |

Taking a content text and append it to the end of 2970 predefined Level 5 SITC categories, we created a corpus *C* consisting of *m* documents and *n* unique terms. The TF-IDF document matrix $M \in R^{m \times n}$ was computed and used as weight vectors. The weighted word embeddings were generated as follows. Every document *t* (such as a cargo content or a predefined category) was padded it to the length n, and represented as a vector $t \in R^{1 \times n}$. Pre-trained Glove embeddings [27] was loaded, and every unique term in the corpus was converted into a $1 \times l$ vector. We used Glove embedding with the word vector dimension of 100, hence *l*=100. We got a word embedding matrix $E \in R^{n \times l}$ to represent the whole corpus. Every document was therefore converted into a matrix of $t = (e_1, \ldots, e_n) \in R^{n \times l}$. Merging the document matrix *t* for the whole corpus *C*, we got a document embedded matrix $DE \in R^{m \times n \times l}$. To calculate pairwise Cosine similarity between the row vectors in $DE \in R^{m \times n \times l}$, we converted every *t*th row $DE_t \in R^{n \times l}$ into a single vector. Calculating weighted average of word vectors in $DE_t$ using the TF-IDF weights, we could reserve the weights by the term uniqueness to the selected content and generate a weighted document embedding matrix $W \in R^{m \times l}$.

Pairwise Cosine similarity was calculated between the cargo content and one of the 2970 predefined SITC categories. After 2970 iterations, the label of the SITC code was assigned as the cluster label to the cargo content, if the Cosine similarity between the instance of cargo content and SITC category is the largest, as expressed in equation (2):

Level 5 label=argmax(Cosine similarity (*cargo content, SITC category*))  (2)

V. TRANSFORMER BASED SUPERVISED CLASSIFICATION

The supervised classification models are built on several Transformers implemented by HuggingFace library [31], including BERT, RoBERTa and XLNet. As shown in Figure 1, Transformer model has two main components, namely the Encoder and the Decoder [16]. The Encoder is composed of 6 identical layers, with each layer having two sub-layers, i.e. a multi-head self-attention mechanism and a simple fully connected Feed Forward network.

The Encoder first performs the encoding of the sentences based on its tokenizer and produces an embedding matrix with an embedding vector for each word. The embedding matrix of the input document is concatenated with a positional encoding; the concatenated matrix enters the multi-head attention block; the multi-head attention layer chooses which parts of the text for the model to focus on; the output from the multi-head attention block x is normalised in a normalisation layer using this function: *LayerNorm(x+Multi-Head Attention(x))*; the normalised output is then fed into a fully connected Feed Forward pass on the neural network and normalised again before being passed to the decoder using a similar function: *LayerNorm(x + FeedForward(x))*.

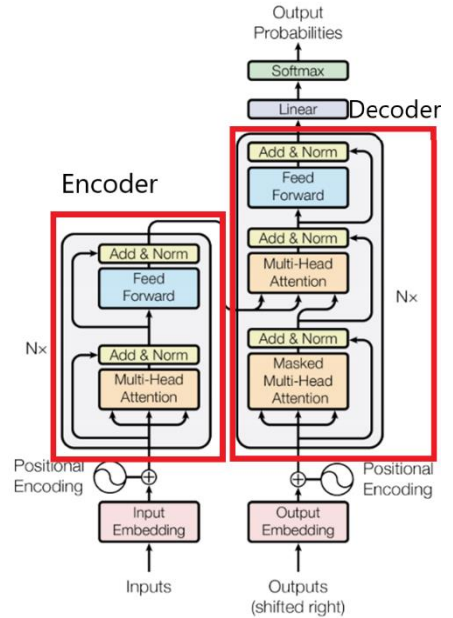

Fig. 1. The encoder-decoder architecture of a Transformer (*adapted from [16]*)

As shown in Fig. 1, the Decoder also has 6 identical layers with each layer having three sub-layers, including the two sub-layers used in the Encoder plus a third sublayer that performs masked multi-head attention directly on the target output sequence. Similar operations as those in the Encoder are employed in each of the three sub-layers. Masked multi-head attention prevents positions from attending to subsequent positions, ensuring that the predictions for one position depends only on the known outputs ahead of this position. The masked multi-head attention is more useful for NLP tasks that predict a sequence of numbers or vectors, such as language translation or next sentence prediction. In text classification, our output is a single output, so the masked multi-head attention sublayer does not add much value. In the Decoder, a FeedForward network receives the final attention vectors and uses them to produce a single vector. Applying the Softmax or other suitable activation functions to this vector produces a set of probabilities belonging to each category.

Three Transformer models were implemented in this research: BERT [32], RoBERTa [33] and XLNet [34]. All the three models have demonstrated high levels of accuracy across NLP tasks, and thus they were selected to classify shipping cargo content when only a small number of training data was available. BERT and RoBERTa are examples of autoencoding based pretraining models where a certain portion of tokens are masked, and the models are trained to recover the original tokens in the context. XLNet is an example of autoregressive language model that predicts a future word given preceding context. BERT is pretrained over hundreds of millions of textual data to learn a language representation that can be fine-tuned for special NLP tasks. Different from Word Embeddings like Glove and Word2Vec that convert the word into one vector, BERT reviews the entire sentence and then assigns an embedding to each word based on the contexts, hence the same word in a different context would be converted to a different vector. The key methods used in BERT include bidirectional Transformers, Masked Language Model and Next Structure Prediction. RoBERTa and XLNet are improved versions of BERT. Both models improved training using much larger data sets and more computational power, leading to better prediction results. Instead of using masked language model, XLNet introduces permutation language modelling where all tokens are predicted in random order. RoBERTa replaces the Next Sentence Prediction task with dynamic masking so that the masked token changes during the training. The set of parameters used in the Transformers models are given in Table II.

TABLE II. PARAMETERS OF TRANSFORMERS

| Transformers | BERT | RoBERTa | XLNet |
|---|---|---|---|
| *Tokenizer parameters* | | | |
| Model name | Bert-base-uncased | Roberta-base | Xlnet-base-cased |
| Max length | 128 | 128 | 128 |
| pad to max length | True | True | True |
| *Transformers parameters* | | | |
| optimizer | AdamW | AdamW | AdamW |
| Loss function | CrossEntropyLoss | CrossEntropyLoss | CrossEntropyLoss |
| Metrics | accuracy | accuracy | accuracy |

VI. TRANING AND TEST DATA

We used real cargo container content to evaluate the different methods described in the previous section. Two sets of training data were constructed: Training Dataset A is composed of 2970 Level 5 SITC textual headings plus 10 Level 1 SITC textual headings, i.e., in total 2980 instances of data. The corresponding numerical codes of these textual headings are used as classification labels (see Appendix 1); Training Dataset B is a bigger set, including Dataset A and additional 970 manually labelled cargo content. The predefined list of headings reported in SITC Level 5 creates a comprehensive training corpus that includes most of the vocabulary relevant to international traded goods. Since the category labels are derived from the Level 1 SITC codes, we merged the headings at SITC Levels 1 and 5 to create Training Dataset A. Two annotators were hired to classify 1000 instances of cargo content using 10 Level 1 codes. Inter-rater reliability based on the percent of agreement between annotators [35] was 84.8% for the entire set. The disagreement between the two annotators was discussed and content was reclassified accordingly. Among the 1000 labelled data, 970 were added to Dataset A to create Dataset B, and the rest 30 instances of manual labelled cargo content data were used to create the test dataset. In the test dataset, each instance has a SITC Level 1 label (0-9) assigned exclusively. Table III presents a description of the datasets.

TABLE III. DATASETS DESCRIPTION

| Dataset | Number of documents | Labels |
|---|---|---|
| *Traning dataset A:* SITC Labels | 2980 | 10 |
| *Traning dataset B:* SITC Labels+manual labelling | 3950 | 10 |
| *Test dataset* | 30 | 10 |

VII. EVALUATION OF CLASSIFICATION PERFORMANCE

The unsupervised and supervised classification models were evaluated by the metric Accuracy (*ACC*). Accuracy is commonly used for comparing multiclass classification [36]. The classification Accuracy (*ACC*) is expressed in terms of the percentage of instances correctly classifies, as shown in equation (3):

$$ACC = C/N \qquad (3)$$

where *C* is the number of stances that are classified correctly, and N is the set of all the testing instances.

To evaluate the actual performance of the unsupervised text classification using Glove word embeddings and Cosine similarity, we compared the unsupervised classification performances with the classification results obtained from a manual labelling process, as shown in Table IV. Since the unsupervised classification used Level 5 labels and the manual labelling process used Level 1 labels, as shown in the columns of "Predicted Level 5 label" and "Manual label" in Table III, the Level 5 labels of the unsupervised classifications were converted to Level 1 labels using the first digit in the Level 5 codes. The "Similarity" column presents the Cosine Similarity of the classification. The Cosine similarity of the classification ranges in [0, 0.99], showing that some content may not be found in the STIC codes while others have good matches.

In Table V, the unsupervised classification results are compared with the performances of the supervised models. Due to the lack of training data, the classification *ACC* of the three Transformer based models are quite low, although feeding additional manually labelled data using Dataset B slightly improved the classification performances. In comparison, it is apparent that the *ACC* of unsupervised classification is much higher. The unsupervised classification model outperforms the three Transformer based supervised classification models, even after the size of the training dataset is significantly increased by 30%. The results show that the unsupervised text classification could act as a practical alternative to categorise cargo content in the absence of labelled training data.

TABLE IV.  EXAMPLES OF EVALUCATING UNSUPERVISED CLASSIFICATION PERFORMANCE

| Cargo content | Predicted Level 5 label | Predicted Level 1 label | Mannual Label | Similarity |
|---|---|---|---|---|
| *Auto parts* | 78439 | 7 | 7 | 0.787 |
| *Baby car seat* | 82112 | 8 | 8 | 0.832 |
| *Electrically Calcined Anthracite Coal* | 32121 | 3 | 3 | 0.844 |
| *Hand sanitiser* | 73515 | 7 | 5 | 0.485 |

TABLE V.  A COMPARISON OF SUPERVISED AND UNSUPERVISED CLASSIFICATION

| Classifiers | Accuracy (ACC) | Increment of Accuracy |
|---|---|---|
| Unsupervised classification | 83% | |
| *Training Dataset A (2980)* | | |
| BERT | 6/30=20% | 0 |
| RoBERTa | 8/30=27% | 0 |
| XLNET | 7/30=23% | 0 |
| *Training Dataset B (3950)* | | |
| BERT | 11/30=37% | 17% |
| RoBERTa | 12/30=40% | 13% |
| XLNET | 10/30=33% | 10% |

The results in Tables IV and V show that the unsupervised classification can produce good results. Further experiments were conducted to classify the 98,321 instances of cargo content using the unsupervised model, and the achieved classification accuracy is 82%. It shows the unsupervised classification model can be applied to deal with large scale text classification problems in real business context, albeit lacking labelled data.

VIII. CONCLUSION, LIMITATIONS AND FUTURE WORK

The unsupervised classification model is built on pre-trained Word Embeddings and Cosine Similarity. In the absence of training data, the efficiency and effectiveness of this model are verified in a real business case where the model is applied to classify shipping cargo content.

Due to the scarce training data, we applied the new state-of-the art Transformers to conduct self-learning and supervised classification. However, the accuracies achieved by the supervised classification models are significantly lower than the ones achieved by the unsupervised classification model. We tested the Transformers based classification across two different training datasets to illustrate the impacts of additional good quality training data. With 30% of additional training data (970 in this research), we can improve the classification accuracy by 10% or more.

While Transformers can potentially reduce the amount of labelled data necessary for achieving optimal levels of classification accuracy, our findings suggest that good performances of these model still depend on high quality and sufficient labelled data.

In the future studies, the unsupervised classification model could be used as a baseline to generate more labelled data for the supervised models. Given that SITC has a large number of closely related categories present in a hierarchical structure, hierarchical multi class text classification will be developed and tested. Hybrid embedding based text representation of all categories in the 5-level SITC hierarchy will be developed, with an aim to improve classification accuracy in the absence of large amount of training data. The hybrid embedding consists of graph embedding of categories in the hierarchy and their word embedding of category labels.